\documentclass[sn-basic]{sn-jnl}

\usepackage{graphicx}%
\usepackage{multirow}%
\usepackage{amsmath,amssymb,amsfonts}%
\usepackage{amsthm}%
\usepackage{mathrsfs}%
\usepackage[title]{appendix}%
\usepackage{xcolor}%
\usepackage{textcomp}%
\usepackage{manyfoot}%
\usepackage{booktabs}%
\usepackage{algorithm}%
\usepackage{algorithmicx}%
\usepackage{algpseudocode}%
\usepackage{listings}%

\theoremstyle{thmstyleone}%
\theoremstyle{thmstyletwo}%

\theoremstyle{thmstylethree}%

\begin{document}

\title[Article Title]{Relation Modeling and Distillation for Learning with Noisy Labels}


\author[1]{Xiaming Chen}
\equalcont{These authors contributed equally to this work.}
\author[1]{\fnm{Junlin} \sur{Zhang}}
\equalcont{These authors contributed equally to this work.}

\author*[2]{\fnm{Zhuang} \sur{Qi}}\email{z\_qi@mail.sdu.edu.cn}

\author*[3]{\fnm{Xin} \sur{Qi}}\email{qixin@usts.edu.cn}

\affil[1]{School of Mathematics and Computer, Shantou University, China.}

\affil[2]{School of Software, Shandong University, China}

\affil[3]{School of Chemistry and Life Sciences, Suzhou University of Science and Technology, China}


\abstract{Learning with noisy labels has become an effective strategy for enhancing the robustness of models, which enables models to better tolerate inaccurate data. Existing methods either focus on optimizing the loss function to mitigate the interference from noise, or design procedures to detect potential noise and correct errors. However, their effectiveness is often compromised in representation learning due to the dilemma where models overfit to noisy labels. To address this issue, this paper proposes a relation modeling and distillation framework that models inter-sample relationships via self-supervised learning and employs knowledge distillation to enhance understanding of latent associations, which mitigate the impact of noisy labels. Specifically, the proposed method, termed RMDNet, includes two main modules, where the relation modeling (RM) module implements the contrastive learning technique to learn representations of all data, an unsupervised approach that effectively eliminates the interference of noisy tags on feature extraction. The relation-guided representation learning (RGRL) module utilizes inter-sample relation learned from the RM module to calibrate the representation distribution for noisy samples, which is capable of improving the generalization of the model in the inference phase. Notably, the proposed RMDNet is a plug-and-play framework that can integrate multiple methods to its advantage. Extensive experiments were conducted on two datasets, including performance comparison, ablation study, in-depth analysis and case study. The results show that RMDNet can learn discriminative representations for noisy data, which results in superior performance than the existing methods.}




\maketitle

\section{Introduction}\label{sec1}

Deep learning techniques have been widely used in the field of computer vision and have made remarkable achievements, such as in object detection \cite{dang2024multi,xia2024human,yang2022focal}, visual tracking \cite{9293326,9344696}, text matching \cite{gan2023tbnf,9474959,9220767}, image recognition \cite{li2024mask,naseem2023approach,9222066}, cross-modal learning \cite{yang2024cross}, \cite{wang2019learning} and other breakthroughs in the field. However, the foundation of these successes is built on accurately labeled datasets. Due to the powerful fitting ability of deep learning models, if noisy labels are present in the dataset, the model may overfit these mislabels during training, which can seriously affect its performance \cite{song2022learning,smart2023bootstrapping,wei2023fine}. 

Many datasets are generated by automatically scraping images and their labels from websites, a method that can introduce noisy labels \cite{lee2018cleannet}, \cite{li2022selective}. While manual annotation can reduce the number of noisy labels, this process is too time-consuming, especially when dealing with large datasets, and manual labeling is not error-free. Studies show that the percentage of noisy labels in actual datasets ranges from 8.0\% to 38.5\% \cite{xiao2015learning,li2017webvision,song2019selfie}. Therefore, noisy annotation is a non-negligible issue that directly affects the effectiveness of model training. To reduce the impact of noisy labels on model training, existing methods mainly fall into two categories: optimization based on the loss function and detection of noisy labels. The former approaches focus on optimizing the existing loss functions to make them more robust, thereby reducing the impact of noisy labels on training \cite{feng2021can,wang2019symmetric,ghosh2017robust,zhang2018generalized}. The latter methods reduces the number of samples with noisy labels by filtering or correcting suspicious labels, thus mitigating the impact of noisy labels \cite{tanaka2018joint,smart2023bootstrapping,kim2019nlnl,nguyen2019self,han2019deep}. However, their performance is typically limited by poor representation learning, which can guide the model to misunderstand during the inference stage.

\begin{figure*}[t]
\centering
\includegraphics[width=0.8\textwidth]{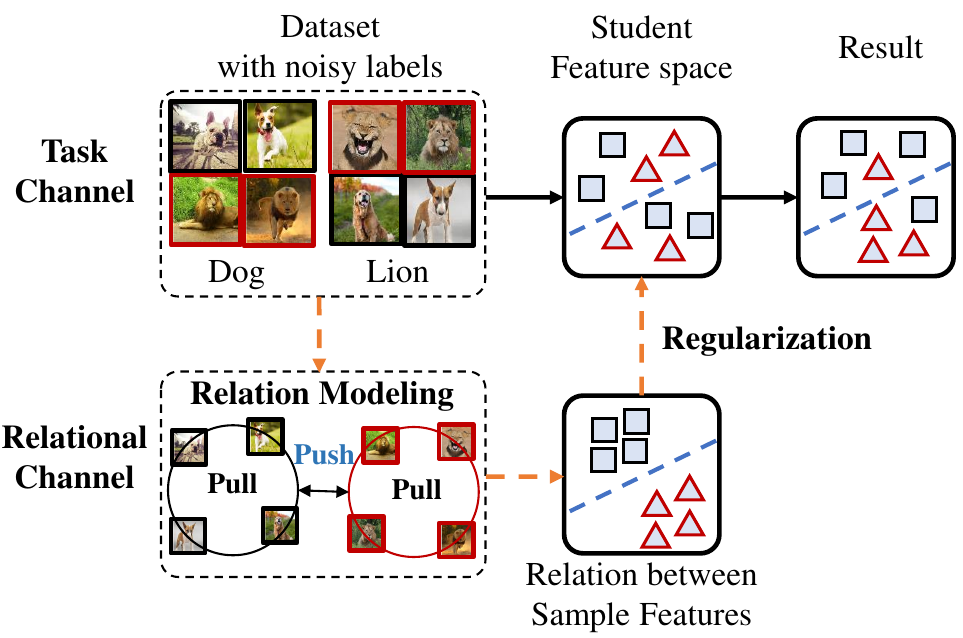}
\caption{Motivation of the proposed RMDNet. Noisy labels diminish the effectiveness of representation learning during the training process. RMDNet firstly models the relation between the representations of samples without the need for label guidance. Secondly, RMDNet utilizes knowledge distillation to regularize representation learning to calibrate the distribution of representations for samples with noisy labels.}
\label{fig1}
\end{figure*}

To address this problem, this paper proposes a relation modeling and distillation framework, termed RMDNet, that mitigates the negative impact of noisy labels by guiding representation learning through modeling the relationships between samples. Fig. \ref{fig1} illustrates its main idea. Specifically, the proposed RMDNet contains two main modules, including the relation modeling (RM) module and the relation-guided representation learning (RGRL) module. The RM module employs self-supervised learning techniques to learn the underlying patterns in data without relying on externally annotated labels. It can extract valuable information and model relationships between samples, mitigating the interference from noise. The RGRL module leverages the knowledge of sample relations acquired from the RM module to guide the learning of data representations, effectively correcting the representation distribution of samples affected by noisy labels. This can improve the robustness of the model during the inference stage. Notably, RMDNet is a general framework that can integrate various methods and enhance their performance.

Extensive experiments were conducted on two datasets, including performance comparison, ablation studies of two main modules, in-depth analysis of key parameters and case studies. The experimental results show that the proposed framework (RMDNet) can improve the robustness of the model in the inference stage by calibrating the representation distribution of samples with noisy labels. 

Overall, the main contributions of the proposed framework can be summarized into three points:

\begin{itemize}
    \item This paper proposes a framework (RMDNet) for learning with noisy labels that models the relation between samples and follows knowledge distillation to mitigate the negative impacts of noisy labels, thereby enhancing the model's robustness.
    \item The proposed RMDNet is an orthogonal improvement to many methods, which can be easily integrated into other methods to benefit them, without needing to change their core architecture. 
    \item This study reveals a finding: errors in representation learning are one of the primary factors that diminish the robustness of models. This ensures the effectiveness of the proposed framework, as RMDNet utilizes relationship modeling and distillation to mitigate the interference of noisy labels on representation learning.
\end{itemize}

The rest of the paper is organized as follows. Section 2 reviews the ideas of learning with noisy labels, contrastive learning techniques, and knowledge distillation methods. In Section 3, we introduce the details of the proposed framework, RMDNet, which includes two modules: the Relationship Modeling module and the Relationship-Guided Representation Learning module. Section 4 presents extensive experiments conducted to evaluate the performance of RMDNet and compares it with other methods. In Section 5, we summarize the RMDNet approach and experimental results and provide an outlook on future work.

\section{Related work}\label{sec2}
\subsection{Noisy Label Learning}\label{subsec2}

In recent years, numerous studies have proposed methods for learning with noisy labels to reduce the interference of noisy labels on model training.
These methods can be broadly categorized into three groups: The first type involves utilizing robust loss functions, which aim to diminish the adverse effects of noisy labels by employing specially designed loss functions. For instance, Taylor Cross Entropy (TCE) \cite{feng2021can} adjusts the fitting of training labels by controlling the order of Taylor series of Categorical Cross Entropy (CCE). Inspired by the symmetric KL scattering, Reverse Cross Entropy (RCE) was used to optimise Cross Entropy (CE) in \cite{wang2019symmetric}, which in turn led to the Symmetric Cross Entropy (SCE). This category also includes losses such as Mean Absolute Error (MAE)  \cite{ghosh2017robust} and Generalized Cross Entropy (GCE) \cite{zhang2018generalized}. The second type aims to reduce the influence of noisy labels by filtering or correcting them. This includes jointly optimizing network parameters and noisy labels to rectify erroneous labels, filtering noise through methods like Negative Learning (NL) \cite{kim2019nlnl}, identifying potential noisy labels during training to prevent their interference with network learning \cite{nguyen2019self}, and employing prototype correction of noisy labels using the self-paced learning framework SMP \cite{han2019deep}. Additionally, some methods utilize re-labeled samples in the final stages of training for classifier training, analyzing the relationship between images and their clean and noisy labels \cite{smart2023bootstrapping}.

\subsection{Contrastive Learning}\label{subsec2}

Contrastive learning, as a form of self-supervised learning method, effectively brings similar samples closer together in the representation space and separates dissimilar samples by using carefully designed models and contrastive loss functions\cite{he2020momentum,chen2020simple,chen2020improved,chen2020big,chen2021empirical,caron2021emerging}. This principle of similarity-based learning has significantly enhanced the performance of representation learning. In recent years, contrastive learning has also been applied in the field of noisy label learning to better address the negative impact caused by noisy labels. For example, the Sel-CL method \cite{li2022selective} learns by identifying confidence pairs within noisy labels and performing contrastive learning on these pairs. Additionally, to obtain high-quality representations from datasets containing noisy labels, researchers have proposed new regularization functions for contrastive learning  \cite{yi2022learning}. Furthermore, twin contrastive learning (TCL) based on Gaussian mixture models \cite{huang2023twin} and UNICON \cite{karim2022unicon} utilize contrastive learning for sample selection, among other techniques. In this study, this paper utilizes contrastive learning to extract representations of dataset, leveraging its self-supervised nature to ensure that noisy labels in the dataset do not affect the learning process, thereby making the extracted representations more reliable.

\subsection{Knowledge Distillation}\label{subsec2}

Knowledge distillation aims to train smaller, lightweight models (student networks) by leveraging supervised information from larger, high-performance models (teacher networks) to enhance their performance and accuracy. In this process, the supervised information obtained from the teacher network's outputs is referred to as "knowledge," and the process of transferring this knowledge to the student network is termed "distillation." For instance, Embedded Graph Alignment (EGA)  \cite{ma2022distilling} transfers knowledge from the teacher's network to the student's network by aligning the student's graph with the teacher's graph. A simple K-way projection method was used in \cite{chen2023accelerating} to transfer knowledge from various levels in the teacher's network to the student's network, distilling knowledge in this way. In DIST \cite{huang2022knowledge}, it was pointed out that the prediction gap between teacher and student networks may be large, and the use of KL scatter will lead to a decrease in the training effect, and the method replaces KL scatter with Pearson correlation coefficient, which achieves some results. Knowledge transfer in CIRKD \cite{yang2022cross} using pixel-to-pixel and pixel-to-region distillation for matching with memory banks.

\section{Method}\label{sec3}
\subsection{Overall framework}\label{subsec2}

\begin{figure*}[t]
\centering
\includegraphics[width=0.85\textwidth]{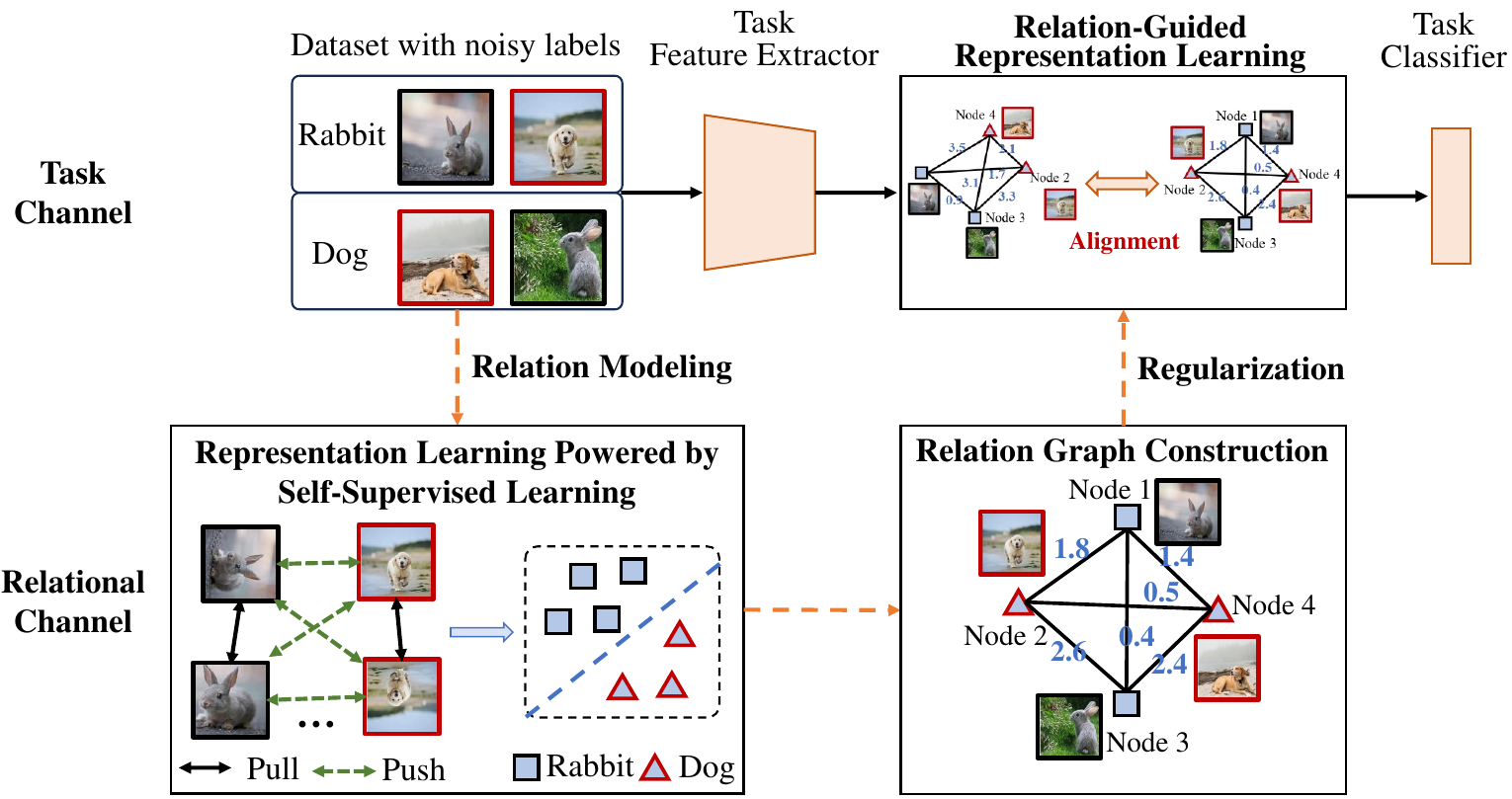}
\caption{The framework of the proposed RMDNet. It contains two main modules, where the Relation Modeling module employs self-supervised learning method to learn discriminative data representations. This can alleviate the interference of noise. The Relation-Guided
Representation Learning module follows knowledge distillation to calibrate the representation distribution with noisy labels.}
\label{fig2}
\end{figure*}

Noisy labels typically introduce incorrect information, causing models to fail to accurately capture the features and patterns of the data, which can results in misinterpretations in the inference stage. To cope with this challenge, this paper proposes a novel RMDNet framework that consists of two key modules, namely the relation modeling (RM) module and the relation-guided representation learning (RGRL) module. Fig. \ref{fig2} illustrates the main idea of the proposed RMDNet. First, the RM module employs self-supervised learning techniques to efficiently extract relational knowledge about the distribution of representations that is not disturbed by noise, ensuring the quality and reliability of the obtained sample representations. Secondly, the RGRL module utilizes the relational knowledge learned from the previous module to mitigate the adverse effects of noisy labels on model training. The next section will describe both modules in detail.

\subsection{Relation Modeling Module}\label{subsec2}
Noise labels often pose significant challenges to model training, such as poor representation learning. To alleviate this issue, the relation modeling (RM) module employs a self-supervised learning method to extract the distribution of representations between classes, which does not rely on the labeling information of the data. Therefore, even in the case of noisy labels, the extracted representational distributions are not affected by the noisy labels.

\subsubsection{Representation learning powered by self-supervised learning}\label{subsubsec2}

\begin{figure*}[t]
\centering
\includegraphics[width=0.5\textwidth]{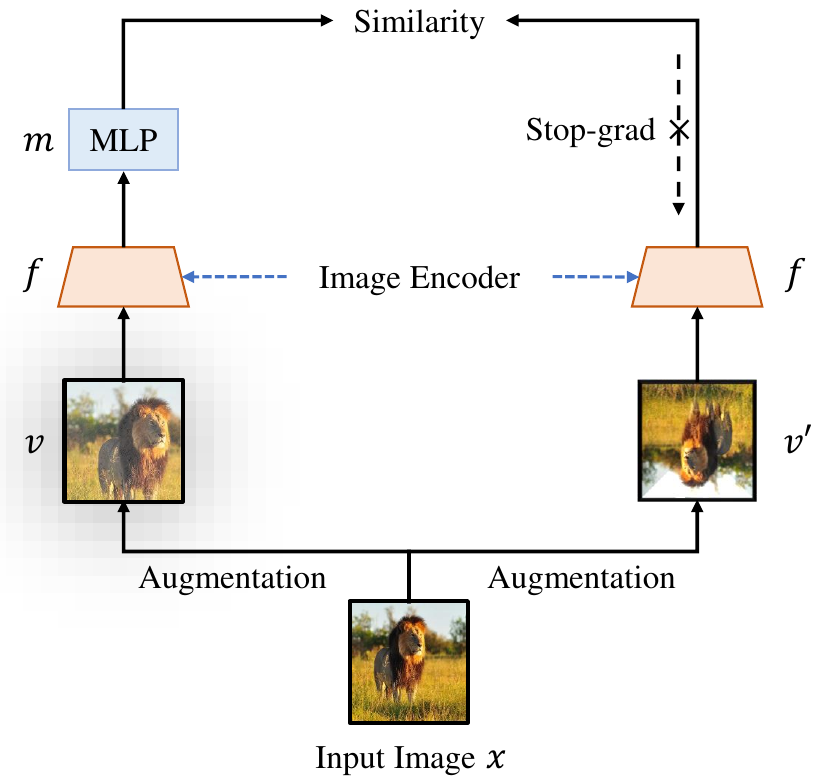}
\caption{The framework of Simsiam.}
\label{fig3}
\end{figure*}

Inspired by contrastive learning, the RM module leverages the SimSiam framework \cite{chen2021exploring} to learn representations that are independent of labels. Its main idea is to maximize the similarity between two views of an image and not to use negative pairs as well as momentum encoders, and also to use stop-gradient to prevent model collapse. As shown in Fig. \ref{fig3}, SimSiam takes as input two augmented views ${v}$ and ${v'}$ of the image ${x}$, which are processed by an identical encoder ${f}$. After passing through an encoder it also passes through an MLP layer ${m}$. The output of ${v}$ after the encoder ${f}$ and the MLP layer ${m}$ is matched with the output of ${v'}$ after the encoder, and the negative cosine similarity of these two vectors is calculated and minimized. If we denote the two outputs as ${q \stackrel{\triangle}{=}m(f(v))}$ and ${y' \stackrel{\triangle}{=}f(v')}$, we have Equation \eqref{eq1}.

\begin{equation}
\label{eq1}
C(q,y')=-\frac {q}{\|q\|_2} \cdot \frac {y'}{\|y'\|_2}
\end{equation}
where is ${\| \cdot \|_2}$ the L2 paradigm. To calculate the symmetric loss, the loss function can be defined as Equation \eqref{eq2}.
\begin{equation}
\label{eq2}
L=\frac {1}{2}C(q,y') + \frac {1}{2}C(q',y) 
\end{equation}

In addition, because of the existence of the operation of stopping the gradient in simsiam, Equation \eqref{eq2} was eventually modified into Equation \eqref{eq3}.
\begin{equation}
\label{eq3}
L=\frac {1}{2}C(q,stopgrad(y')) + \frac {1}{2}C(q',stopgrad(y)) 
\end{equation}
where $stopgrad(\cdot)$ denotes the stop gradient. And this method can learn discriminative representations without the need for label guidance, mitigating the interference of noisy labels and providing reliable information for relation graph construction.

\subsubsection{Relation Graph Construction}\label{subsubsec2}

To obtain relational knowledge in the representation space, the method encodes the correlation between a group of representations by constructing a graph $G=\{(N,E)\}$, where $\{N\}$ denotes a set of nodes and $\{E\}$ denotes a set of edges to encode the correlation between different samples, note ${E \subseteq \{(x_i,x_j)|(x_i,x_j) \in N \enspace  and \enspace  x_i \neq x_j \} }$.

To extract the knowledge of the relation of representation distribution between different samples, the method encodes the correlation between each pair of samples in the same batch by deriving each edge in the graph using Pearson correlation coefficient (PPC). Now suppose there are m pairs of samples, denoted ${ \{ (x_1,y_1),(x_2,y_2), \cdots ,(x_m,y_m) \} }$, then the PPC can be defined as Equation \eqref{eq4}.
\begin{equation}
\begin{aligned}
\label{eq4}
r_{xy} &=  \frac {\sum_{i=1}^{m} {(x_i- \overline{x})(y_i- \overline{y})}}{(m-1)S_xS_y} \\
       &=  \frac {\sum_{i=1}^{m} {(x_i- \overline{x})(y_i- \overline{y})}}{\sqrt[]{\sum_{i=1}^{m} {(x_i- \overline{x})^2}}
         \sqrt[]{\sum_{i=1}^{m} {(y_i- \overline{y})^2}}}
\end{aligned}
\end{equation}
where ${m}$ denotes the sample size, $({x_i},{y_i})$ denote a data point, ${\overline{x}}$ and ${\overline{y}}$ denote the mean value, and ${S_x}$ and ${S_y}$ denote the standard deviation.

The correlation between the representations of each pair of data nodes is quantified using Equation \eqref{eq4}, and the edges are also constructed using this correlation. Specifically, given a pair of sample representations, we denote them as ${x}$ and ${y}$ (where, ${x \in R^D}$, ${y \in R^D}$).

According to the formula Equation \eqref{eq4}, we can calculate the edges between these data. As shown in equation Equation \eqref{eq5}.
\begin{equation}
\label{eq5}
e_{x,y}=\frac {\sum_{i=1}^{D} {(x_i- \overline{x})(y_i- \overline{y})}}
        {\sqrt[]{\sum_{i=1}^{D} {(x_i- \overline{x})^2}}
         \sqrt[]{\sum_{i=1}^{D} {(y_i- \overline{y})^2}}} 
\end{equation}
where ${D}$ denotes the dimension of the feature, and ${e_{x,y}}$ represents the edges between the representations of two samples ${x}$ and ${y}$ and quantifies the relationship between them by their correlation. By using Equation \eqref{eq5}, the edges between the representations of each pair of samples in a batch of node representations can be calculated, and such a process is also used to construct the student graph on the task channel.

\subsection{Relation-Guided Representation Learning Module}\label{subsec2}

\begin{figure*}[t]
\centering
\includegraphics[width=0.6\textwidth]{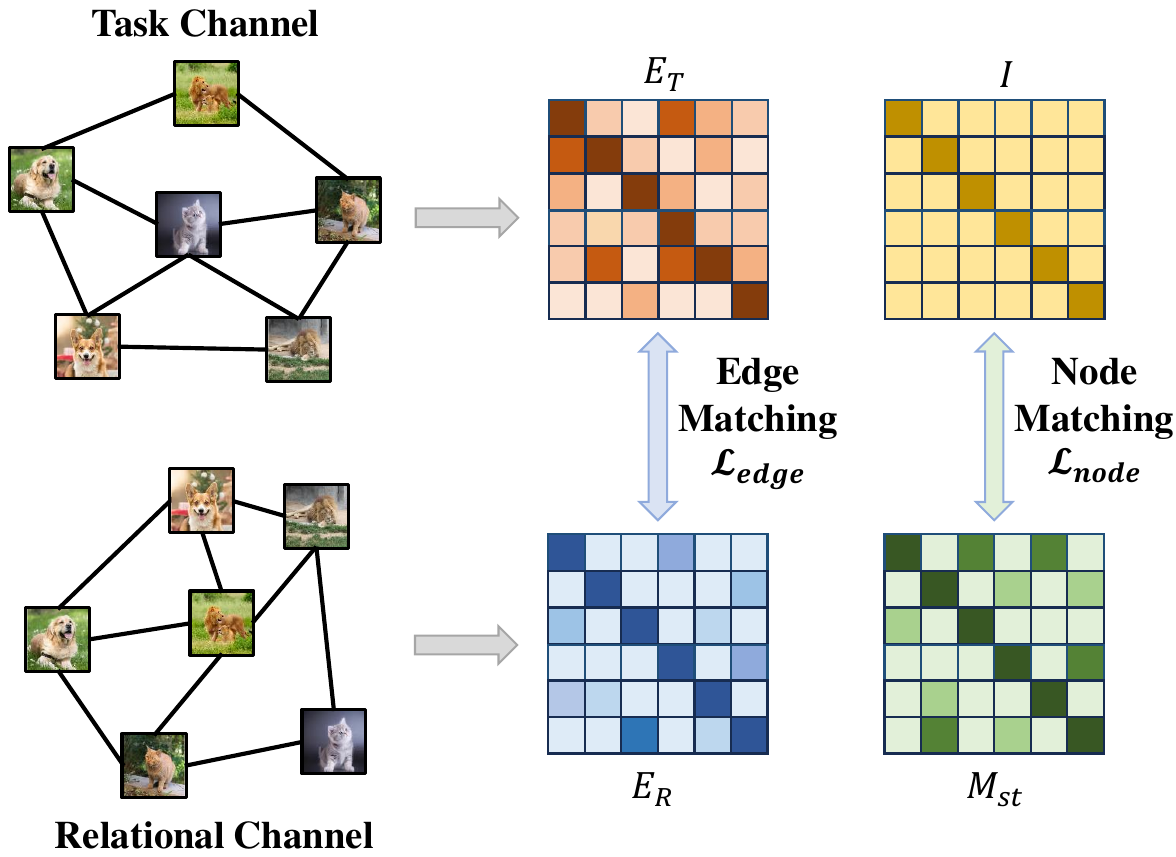}
\caption{Illustration of the Relation-Guided Representation Learning method. It contains two main constraint conditions, including the edge matching and the node matching.}
\label{fig4}
\end{figure*}

To alleviate the interference of noise labels on model training, the relation-guided representation learning (RGRL) module employs knowledge distillation method to transfer pre-trained relation graph to model training. Specifically, in the task channel, noisy data is used as input; in the relational channel, the pre-trained relation graph is loaded, and serves as a regularizer to guide the representation learning of the task channel, which can improve the robustness of the model in the inference phase. As shown in Fig. \ref{fig4}, the basic idea of this method is to transfer inter-instances relational knowledge from the relational channel to the task channel by aligning the teacher and student graphs through edge matching constraints and node matching constraints.

The following briefly describes how to construct edge matching constraints and node matching constraints via edge matrix and node matrix respectively, and how they can be used for knowledge transfer.

\subsubsection{Edge Matrix Calculation}\label{subsubsec2}

Now encode a batch of samples representing all pairwise correlations between ${N=\{x_1,x_2, \cdots ,x_B\}}$. Suppose ${e_{x,y}}$ is the correlation between representation ${x_i}$ and representation ${x_j}$ computed according to Equation \eqref{eq5}. Then the edge matrix of this batch of sample representations can be expressed as Equation \eqref{eq6}.
\begin{equation}
\label{eq6}
E(N,N)=(e_{i,j}) \in R^{B \times B}
\end{equation}
where ${B}$ denotes the number of samples in each batch of data, the elements on the diagonal of ${E}$ denote the correlation of each representation to the representation itself, and therefore these elements have the value of 1. The node representations of the relational and task channels can be calculated by using the Equation \eqref{eq6} to calculate their corresponding edge matrices. Assume that the node representations of the two network outputs are ${N=\{x_{t_1},x_{t_2}, \cdots ,x_{t_B}\}}$ and ${N=\{x_{s_1},x_{s_2}, \cdots ,x_{s_B}\}}$ respectively. The corresponding edge matrices can be written as ${E_t=E(N_t,N_t)}$ and ${E_s=E(N_s,N_s)}$, respectively.

\subsubsection{Edge Matching}\label{subsubsec2}

In order to learn relational knowledge from the relational channel, it is necessary to align the edge matrices of the teacher graph of relational channel with the edge matrices of the student graph of the task channel. This can be achieved by using edge matching loss as shown in Equation \eqref{eq7}.

\begin{equation}
\label{eq7}
\mathcal{L}_{edge} \stackrel{\triangle}{=} \|E_t-E_s\|_2
\end{equation}
where ${\mathcal{L}_{edge}}$ forces the task channel to extract the same relational knowledge as the relational channel by aligning the edge matrices ${E_t}$ and ${E_s}$. By training with such constraints, the task channel obtains similar relational knowledge among different instances in the representation space.

\subsubsection{Node Matrix Calculation}\label{subsubsec2}

Although Equation \eqref{eq7} facilitates the transfer of relational knowledge from the relational channel to the task channel, the individual representations between the two channels may not be aligned. Therefore, in order for the task channel to better learn the representations learned from the relational channel, the representations of the same sample need to be aligned. This can be achieved by defining a node matrix in which each element is used to capture the correlation between the output representations of the two channels. The node matrix is derived in a similar way to Equation \eqref{eq6}. This can be expressed as the Equation \eqref{eq8},

\begin{equation}
\label{eq8}
M_{st} = E(N_t,N_s)
\end{equation}
where ${M_{st}}$ denotes the node matrix, the interrelationships between the relational and task channels are quantified in this way by creating an edge in each pair of relational and task nodes to connect the relational and task representations.

\subsubsection{Node Matching}\label{subsubsec2}

Using node matching loss can align the representations of the relational channel and task channel so that the same input samples are highly correlated in the output representations of both channels, while different samples are less correlated. Node matching can be achieved by enforcing Equation \eqref{eq8} to align with the unit matrix ${I}$, as shown in Equation \eqref{eq9}.

\begin{equation}
\label{eq9}
\mathcal{L}_{node} \stackrel{\triangle}{=} \|M_{st}-I\|_2
\end{equation}
In the formula, try to make the data on the diagonal ${M_{st}}$ 1 and the rest 0. This is because the data on the diagonal represents the correlation between the same sample representations in the relational channel and task channel, and the larger the value the greater the correlation between the two sample representations.

\subsubsection{Alignment of Relation Graphs}\label{subsubsec2}

In order to align the teacher graph of relational channel with the student graph of task channel, the embedded graph alignment loss is used for distillation, which contains two loss terms: the edge matching loss ${\mathcal{L}_{edge}}$ and the node matching ${\mathcal{L}_{node}}$ loss. It is defined as in Equation \eqref{eq10}.
\begin{equation}
\label{eq10}
\mathcal{L}_{RMDNet} = \alpha \mathcal{L}_{node} + \beta \mathcal{L}_{edge}
\end{equation}
where ${\alpha}$ and ${\beta}$ are the weight hyperparameters for edge matching loss and node matching, respectively.

\subsubsection{Overall Optimization of the Model}\label{subsubsec2}

Let the weight hyperparameter of ${L_{RMDNet}}$ be ${K}$. If we use CE loss, the final loss function is shown in Equation \eqref{eq11} as follows.
\begin{equation}
\label{eq11}
\mathcal{L} = \mathcal{L}_{CE} + K \cdot \mathcal{L}_{RMDNet}
\end{equation}

We achieve knowledge transfer by using ${\mathcal{L}_{RMDNet}}$ loss to allow the pre-trained relational knowledge to guide the task network to train, and in this way reduce the effect of noise labeling on the task network and improve the accuracy of model training.

\section{EXPERIMENTS}\label{sec4}

In this section, we describe the experimental design in detail, including the dataset used, the method of noise generation, and other key experimental details. In addition, we demonstrate the effectiveness of our method by comparing it with other existing methods. To further validate the importance of each module, we also conduct a series of ablation experiments. This chapter will also explore the effect of the selection of the hyperparameter K on the experimental results. Finally, we will show in detail the effectiveness of our method in specific cases by visualizing and analyzing the model representations.

\subsection{Datasets}\label{subsec2}

In order to validate the effectiveness of our algorithm, we designed a series of experiments involving two main datasets and three different types of noise conditions, as well as three different noise rates. The details are as follows:

The two datasets are CIFAR-10 and CIFAR-100, both of which are both 32×32×3 color images and include 50,000 training images and 10,000 test images. In CIFAR-10, the images are categorized into 10 classes, while CIFAR-100 categorizes the images into 100 classes.

Three types of synthetic labeling noise, symmetric noise, asymmetric noise, and flip noise, with three noise rates of 10\%, 20\%, and 40\%, respectively. The details of noise production are as follows:

Symmetric noise: this labeling noise is generated by flipping the labels in each class uniformly onto the labels of the other classes to produce noisy labels.

Asymmetric noise: this labeling noise is generated by flipping labels in a set of similar classes. In this paper, CIFAR-10: Truck→Car, Bird→Airplane, Deer→Horse, Cat→Dog, Dog→Cat. CIFAR-100: 100 classes are divided into 20 classes, each class has 5 subclasses, and then each class is flipped to the next class in the same class.

Pairflip noise: the noise flips each class to a neighboring class, for an explanation of this noise see \cite{zheng2020error}, \cite{lyu2019curriculum}.

By conducting experiments under these diverse test conditions, we were able to gain a detailed understanding of the performance and stability of the algorithm in the face of different types and intensities of noise.

\subsection{Evaluation Measures}\label{subsec2}

In this paper, the experiments are evaluated using the accuracy rate, which is given in Equation \eqref{eq12}.
\begin{equation}
\label{eq12}
Accuracy=\frac {TP+TN}{TP+TP+TN+FN} 
\end{equation}
In the formula, TP (True Positive) indicates that positive samples were correctly classified, FP (False Positive) indicates that positive samples were incorrectly classified, FN (False Negative) indicates that negative samples were incorrectly classified, and TN (True Negative) indicates that negative samples were correctly classified. TP+TN denotes the portion of samples which were correctly categorized, TP+TN+FP+FN denotes the entire data, and the accuracy is calculated as the number of correctly categorized samples divided by the total number of samples.

\subsection{Experimental Details}\label{subsec2}

We trained on two datasets, CIFAR-10 and CIFAR-100, using the ResNet18 encoder with the parameters kernel size set to 3, stride set to 1, and padding set to 1 for nn.Conv2d.

In the relation modelling module, we utilize the SimSiam framework for self-supervised learning. To optimize the training process, we chose SGD as the optimizer and set momentum to 0.9 and weight decay to 0.0005. For both CIFAR-10 and CIFAR-100 datasets, the initial learning rate (LR) is set to 0.03. In addition, the batch size is set to 64 and the network training lasts for 1000 epochs, and the learning rate was adjusted using cosine learning rate decay.

In the relation-guided representation learning (RGRL) module, the task channel also use the SGD optimizer with momentum set to 0.9 and weight decay set to 0.0005. for CIFAR-10, the initial learning rate is set to 0.001, while for CIFAR-100, the initial learning rate is 0.01. we train each network for 240 epochs while decaying the learning rate by a factor of 0.1 at the 150th epoch and every 30epoch thereafter. Common data enhancement techniques were applied in the experiments, including random cropping and level flipping. The hyperparameter K takes values ranging from 0.0001 to 2.0, depending on the loss function used, the type of noise, and so on. For the hyperparameters of the ${L_{RMDNet}}$ loss, the node weights ${\alpha}$ and edge weights ${\beta}$ we set to 0.8 and 0.35, respectively.

\begin{table*}[t]
	\centering
	\caption{Experimental results of RMDNet with baseline, on CIFAR-10 and CIFAR-100 datasets, where there are three types of noise symmetric, asymmetric, and flipped noise, with noise rates of 0.1, 0.2, and 0.4, respectively. \label{tab:table1}}
	\vspace{5pt}
    \resizebox{0.89\textwidth}{!}{ 
	\begin{tabular}{ccccccccccc}\hline
		\multirow{2}{*}{\centering Datasets} &\multirow{2}{*}{\centering Methods} &\multicolumn{3}{c}{Symmetric} &\multicolumn{3}{c}{Asymmetric} &\multicolumn{3}{c}{Pairflip} \\
		\cmidrule{3-11}   
		& & 0.1 & 0.2 & 0.4 & 0.1 & 0.2 & 0.4 & 0.1 & 0.2 & 0.4 \\ \cmidrule{1-11} 
            \multirow{10}{*}{\centering CIFAR-10}
		&CE &86.92	&80.73	&62.12	&90.32	&85.79	&76.48	&88.04	&79.46	&56.64\\
		&GCE &91.77	&89.15	&81.47	&91.06	&86.49	&74.27	&90.88	&83.95	&55.18\\
		&SCE &88.61	&82.37	&63.46	&90.63	&85.89	&76.43	&87.59	&79.66	&55.53\\
            &NCE\_AGCE &92.18	&91.07	&87.55	&91.43	&90.34	&82.49	&91.66	&90.18	&80.37\\  
            &NCE\_AEL &90.79	&89.27	&83.66	&91.30	&87.92	&75.33	&90.60	&87.08	&58.71\\ \cmidrule{2-11} 
            &\textbf{CE+RMDNet} &89.63	&85.20	&71.00	&90.97	&87.51	&77.15	&89.23	&82.07	&58.81\\
            &\textbf{GCE+RMDNet} &92.15	&91.89	&85.61	&91.46	&87.81	&76.73	&91.68	&89.15	&57.84\\
            &\textbf{SCE+RMDNet} &90.24	&85.14	&68.91	&91.38	&87.66	&77.13	&89.13	&81.38	&59.35\\
            &\textbf{NCE\_AGCE+RMDNet} &\textbf{93.38}	&91.86	&87.71	&\textbf{92.52}	&\textbf{90.38}	&\textbf{82.99}	&92.43	&90.57	&\textbf{81.55}\\
            &\textbf{NCE\_AEL+RMDNet} &92.27	&\textbf{91.95}	&\textbf{91.01}	&91.89	&90.34	&77.24	&\textbf{92.49}	&\textbf{90.81}	&59.31\\
		\hline
		\multirow{10}{*}{\centering CIFAR-100}
		&CE &68.21	&60.63	&46.31	&69.90	&62.16	&44.39	&68.85	&62.69	&45.14\\
		&GCE &74.23	&72.73	&67.29	&73.23	&69.37	&46.56	&72.77	&69.54	&46.65 \\
		&SCE &66.99	&60.07	&42.78	&67.94	&60.03	&43.44	&67.26	&59.93	&43.10\\
            &NCE\_AGCE &71.53	&68.92	&57.47	&72.38	&69.64	&44.77	&70.51	&69.24	&46.5\\
            &NCE\_AEL &68.84	&61.45	&35.73	&70.35	&61.61	&43.08	&69.36	&61.30	&42.95\\ \cmidrule{2-11} 
            &\textbf{CE+RMDNet} &68.95	&63.48	&52.47	&69.93	&63.79	&45.63	&69.83	&63.53	&45.17\\
            &\textbf{GCE+RMDNet} &\textbf{75.61}	&74.10	&\textbf{68.60}	&\textbf{75.37}	&69.67	&47.12	&\textbf{75.09}	&70.04	&47.71\\
            &\textbf{SCE+RMDNet} &67.81	&61.61	&46.56	&68.96	&61.74	&44.99	&68.68	&61.51	&44.26\\
            &\textbf{NCE\_AGCE+RMDNet}  &74.82	&\textbf{72.55}	&60.69	&75.07	&\textbf{72.79}	&\textbf{48.89}	&75.07	&\textbf{72.81}	&\textbf{47.84}\\
            &\textbf{NCE\_AEL+RMDNet}  &70.01	&62.34	&41.98	&70.68	&61.95	&44.21	&69.99	&61.90	&43.77\\
		\hline
	\end{tabular}}

\end{table*}

\subsection{Performance Comparison}\label{subsec2}

In this study, we apply the RMDNet framework to several state-of-the-art loss function methods, including generalized cross-entropy (GCE)\cite{zhang2018generalized}, symmetric cross-entropy (SCE)\cite{wang2019symmetric}, NCE\_AUL is NCE combined with AUL, NCE\_AGCE is NCE combined with AGCE \cite{zhou2023asymmetric}. In addition, we have the traditional cross-entropy loss (CE).

All of these methods use ResNet18 as the base encoder, a network architecture consisting of a 3x3 convolutional layer, a batch normalization layer and a ReLU activation function, followed by four 3x3 convolutional layers. The network ends with an adaptive average pooling layer and a linear layer. The experimental results are summarized in Table \ref{tab:table1}, which demonstrates the performance of each method under different loss functions, datasets, noise types and noise rates. In these experiments, the hyperparameter K takes values ranging from 0.0001 to 2.0.

\begin{itemize}
\item From the analysis of the data in Table \ref{tab:table1}, it is concluded that the results obtained with RMDNet are higher than the baseline results in both datasets and in the experiments with different combinations of noise types and noise rates, as compared to the other baseline results. This finding confirms that our proposed method enhances the robustness of the loss function and effectively reduces the negative impact of noise on model training.

\item Furthermore, we observe that RMDNet is more effective in improving those loss functions that are less robust. For example, when applied to the CE loss and confronted with symmetric noise and a noise rate of 0.2, the raw accuracy is 80.73\%. With RMDNet, the accuracy is significantly improved to 85.20\%, an increase of 4.47\%. In contrast, for the NCE\_AGCE loss, which is inherently highly robust, the raw accuracy is 91.07\%, also with symmetric noise and a noise rate of 0.2, and improves to 91.86\% with the use of RMDNet, an improvement of only 0.79\%. This shows that RMDNet is more effective for loss functions that are less robust.

\end{itemize}

\subsection{Ablation Experiments}\label{subsec2}

In this section, we delve into the effectiveness of two core modules in the RMDNet framework, which consists of two parts: first, the relation modeling (RM) module for self-supervised learning, the main function of which is to use self-supervised learning to extract relational knowledge about representation distributions that are not affected by noise; and second, the relation-guided representation learning (RGRL) module, this module guides the training of the task network by loading the pre-trained relation graph on the relational channel and applying it to guide the training of the task network and reduce the effect of noisy labels on the task network.

In this section, we delve into the effectiveness of two core modules in the RMDNet framework: first, the Relational Modelling (RM) module, whose main function is to use self-supervised learning to extract noiselessly distributed relational knowledge of the representations; second, the Relationally-Guided Representation Learning (RGRL) module, which reduces the effects of noisy labels on the task network by loading the pre-trained relation graph, and applying the relational knowledge to guide the task network's training.

In this ablation experiment, we use CE loss as the experimental benchmark. In addition, since it is necessary to use the relational knowledge acquired by the RM module in the RGRL module, we are divided into two cases of using the relational knowledge and not using the relational knowledge when performing the ablation experiment. In addition, the datasets used for the experiments include CIFAR-10 and CIFAR-100, respectively, under symmetric noise conditions, with the noise rate set to 0.1 and 0.2, and the K-value set to 1. Comparison experiments with K=5 were conducted to observe the effects of different K-value settings, the experimental results are recorded in Table \ref{tab:table2}.

\begin{table*}[t]
	\centering
	\caption{In order to explore the functionality of the different modules of RMDNet, the results obtained by using the different modules of RMDNet acting on CE and training on CIFAR-10 and CIFAR-100 datasets are now used. \label{tab:table2}}
	\vspace{5pt}
	\begin{tabular}{cccc}\hline
		\multirow{2}{*}{\centering Datasets} &\multirow{2}{*}{\centering Methods}  &\multicolumn{2}{c}{Symmetric} \\
		\cmidrule{3-4}   
		& & 0.1 & 0.2  \\ \cmidrule{1-4} 
            \multirow{4}{*}{\centering CIFAR-10}
		&CE	&86.92	&80.73	\\
		&CE+RGRL	&85.85	&79.31	\\
		&CE+RGRL (K=5)	&82.52	&76.91	\\
            &\textbf{CE+RGRL+RM}	&\textbf{89.63}	&\textbf{85.20}	\\  \hline
		\multirow{4}{*}{\centering CIFAR-100}
		&CE	&68.21	&60.63	\\
		&CE+RGRL	&64.56	&57.40	 \\
		&CE+RGRL (K=5)	&60.10	&50.62	\\
            &\textbf{CE+RGRL+RM}	&\textbf{68.95}	&\textbf{63.48}	\\
		\hline
	\end{tabular}
\end{table*}

Based on the analysis results in Table \ref{tab:table2}, we find that adding RGRL only on CE loss leads to a decrease in the accuracy of model training. This phenomenon is mainly caused by two factors:

\begin{itemize}
\item Firstly, the teacher network in the relational channel, as a static network, does not involve any optimizer during the training process, so its parameters remain constant. 

\item Secondly, the teacher network did not use the relational knowledge gained from the RM module, so its representation learning was consistently at a poor level throughout the training process. If such a teacher network is relied upon to guide the task network, the result naturally leads to a decrease in accuracy.
\end{itemize}

Specifically, for example, in the CIFAR-10 dataset, when the noise rate is 0.1 and 0.2, the accuracy rate drops by 1.07\% and 1.42\%, respectively, compared with the baseline; in CIFAR-100, the drop in the accuracy rate is even more pronounced, 3.65\% and 3.23\%, respectively. It is worth noting that the decrease in accuracy is even greater when the K value is increased, for example, when the K value is set to 5 and the noise rate is 0.1 and 0.2 in the CIFAR-10 dataset, the accuracy is lower than the baseline by 4.40\% and 3.82\%, respectively; while in the CIFAR-100 dataset the accuracy is lower than the baseline by 8.11\% and 10.01\%, respectively.

However, the situation improves if the teacher network in the RGRL module uses the relational knowledge acquired by the RM module. Since the RM module is trained by self-supervised learning and is not affected by noisy labels, its extracted representation is more reliable. Using this reliable representation to guide the task network can significantly improve the accuracy of the training.For example, under the CE+RGRL+RM configuration, the CIFAR-10 dataset shows an accuracy improvement of 2.71\% and 4.47\% compared to the baseline at noise rates of 0.1 and 0.2, respectively; in CIFAR-100, the improvement is 0.74\% and 2.85\%, respectively.

To summarize, the two modules in our approach, both of which play their respective roles, will work better when combined together.

\subsection{Analysis of Key Parameters}\label{subsec2}
In this section, we will delve into the specific impact of the hyperparameter K value on model training. It is important to note that different loss functions, noise types, noise rates, and datasets may affect the setting of the K-value. In order to analyze these effects in detail, we chose two loss function combinations, NCE\_AGCE+RMDNet and CE+RMDNet, for our experiments, and two datasets, CIFAR-10 and CIFAR-100, were also used.

It is worth emphasizing that the K-values used in this section are not necessarily the optimal K-values, and the optimal K-values mentioned here are those that can achieve the highest accuracy. The experimental results are presented in Table \ref{tab:table3}, which shows the training results with different K-values on the CIFAR-100 dataset for the NCE\_AGCE+RMDNet and CE+RMDNet loss functions when the noise rate is 0.2 and 0.4. In particular, when the K value is set to 0, it indicates that our RMDNet method is not applied, and the results at this point represent the baseline accuracy.

\begin{table*}[!htbp]
	\centering
	\caption{Experimental results of NCE\_AGCE+RMDNet, CE+RMDNet on CIFAR-100 with symmetric noise rate of 0.2 and 0.4. \label{tab:table3}}
	\vspace{5pt}
	\begin{tabular}{cccc}\hline
		\multirow{2}{*}{\centering Loss} &\multirow{2}{*}{\centering K}  &\multicolumn{2}{c}{Symmetric} \\
		\cmidrule{3-4}   
		& & 0.2 & 0.4  \\ \cmidrule{1-4} 
            \multirow{3}{*}{\centering CE+RMDNet}
		&1	&\textbf{63.48}	&\textbf{52.47}	\\
		&0.001	&61.05	&45.72	\\
            &0	&60.63	&46.31	\\  \hline
		\multirow{3}{*}{\centering NCE\_AGCE+RMDNet}
		&1	&1.00	&1.00	\\
		&0.001	&\textbf{69.47}	&\textbf{60.69}	 \\
		&0	&68.92	&57.47	\\
		\hline
	\end{tabular}
\end{table*}

\subsubsection{The effect of the loss function on the value of K}\label{subsubsec2}

As the results presented in Table \ref{tab:table3}, using the CE+RMDNet loss function combination performs well when the value of K is set to 1, improving the accuracy by 2.85\% and 6.16\%, respectively, compared to the baseline accuracy. On the contrary, for the NCE\_AGCE+RMDNet combination performs poorly at the same setting. However, when the value of K decreases to 0.001, NCE\_AGCE+RMDNet not only outperforms the baseline, but also exceeds the effect of the combination of CE+RMDNet and NCE\_AGCE. This suggests that different loss functions have different sensitivities to the K value, and adjusting the K value can affect the results of training.

\subsubsection{Effect of noise type on K-value}\label{subsubsec2}

Table \ref{tab:table4} shows the experimental results using the NCE\_AGCE+RMDNet loss function on the CIFAR-10 dataset for three noise types (symmetric, asymmetric, and flipped) and two noise rates (0.2 and 0.4). When the value of K is 0, the RMDNet method is not applied and reflects the baseline accuracy. From the data, it can be seen that at a noise rate of 0.2, the setting of K=1 under symmetric noise resulted in a 0.79\% improvement in accuracy over the baseline, while the accuracy for asymmetric and flipped noise was below the baseline by 0.95\% and 3.66\%, respectively. When the value of K is reduced to 0.001, the accuracy of symmetric and asymmetric noise is 0.21\% and 0.09\% lower than the baseline, while the accuracy of flip noise is 0.67\% higher than the baseline. This indicates that the value of K may be different for better training results under different noise types.

\subsubsection{Effect of noise rate on K-value}\label{subsubsec2}

In addition, by comparing the results for a value of K of 1, symmetric noise and a noise rate of 0.2 versus 0.4, it was able to conclude that the accuracy was 0.79\% higher than the baseline at the low noise rate, while the accuracy was 1.10\% lower than the baseline at the high noise rate. At the setting of K=0.001, although the accuracy is lower than the baseline for both noise rates, the accuracy is 0.56\% higher at 0.4 noise rate than at K=1. This further suggests that a proper adjustment of K may yield superior training results under different noise conditions. In particular, when the noise rate increases, appropriately lowering the K value may lead to better performance.

\begin{table*}[!htbp]
	\centering
	\caption{Experimental results of NCE\_AGCE+RMDNet loss on CIFAR-10 with three noise types and two noise rates with different K values. \label{tab:table4}}
	\vspace{5pt}
	\begin{tabular}{ccccccc}\hline
		\multirow{2}{*}{\centering K}  &\multicolumn{2}{c}{Symmetric} &\multicolumn{2}{c}{Asymmetric} &\multicolumn{2}{c}{Pairflip} \\
		\cmidrule{2-7}   
		&0.2 &0.4  &0.2 &0.4  &0.2 &0.4 \\ \cmidrule{1-7} 
            \multirow{1}{*}{\centering 1}
		&\textbf{91.86}	&86.45 	&89.39	&76.94	&86.52	&58.79\\
            \multirow{1}{*}{\centering 0.001}
            &90.86	&87.01	&90.25	&\textbf{82.68}	&\textbf{90.57}	&\textbf{81.04}\\
            \multirow{1}{*}{\centering 0}
            &91.07	&\textbf{87.55}	&\textbf{90.34}	&82.49	&90.18	&80.37\\
		\hline
	\end{tabular}
\end{table*}

\begin{table*}[!htbp]
	\centering
	\caption{Experimental results of NCE\_AGCE+RMDNet loss on the same K-value and noise rate, different datasets. \label{tab:table5}}
	\vspace{5pt}
	\begin{tabular}{cccc}\hline
		\multirow{2}{*}{\centering Datasets} &\multirow{2}{*}{\centering K}  &\multicolumn{2}{c}{Symmetric} \\
		\cmidrule{3-4}   
		& & 0.2 & 0.4  \\ \cmidrule{1-4} 
            \multirow{3}{*}{\centering CIFAR-10}
		&1	&\textbf{91.86}	&86.45 	\\
		&0.001	&90.86	&87.01	\\
            &0	&91.07	&\textbf{87.55}	\\  \hline
		\multirow{3}{*}{\centering CIFAR-100}
		&1	&1.00	&1.00	\\
		&0.001	&\textbf{69.47}	&\textbf{60.69}	 \\
		&0	&68.92	&57.47	\\
		\hline
	\end{tabular}
\end{table*}

\subsubsection{Effect of data set on K-value}\label{subsubsec2}

In Table \ref{tab:table5}, we show in detail the experimental results using the NCE\_AGCE+RMDNet loss on the CIFAR-10 and CIFAR-100 datasets with symmetric noise, and noise rates of 0.2 and 0.4. Table \ref{tab:table5} clearly shows the significant effect of K value on the training results. On the CIFAR-10 dataset, when the K value is set to 1, good training results are obtained. However, the same setting on the CIFAR-100 dataset fails to achieve effective training with very low accuracy. On the contrary, when the K value is adjusted to 0.001, it performs better on CIFAR-100, and the accuracy is improved by 0.55\% and 2.22\% under the conditions of symmetric noise rate of 0.2 and 0.4, respectively. This result suggests that the K value needs to be set differently on different datasets in order to achieve better training results.

In summary, the optimal setting of the K-value is affected by a variety of factors, including different loss functions, noise types, noise rates, and data sets. Therefore, the optimal setting of the K-value needs to be determined by continuous adjustment and testing during the experimental process. Our suggestion is to first perform a wide range of K-value adjustments to observe the change in accuracy; when the accuracy does not change much, then make fine adjustments around that value to fine-tune the determination of the optimal K-value.

\subsection{Case Studies}\label{subsec2}

In this section, we demonstrate the effectiveness of the RMDNet method by visualizing the extracted representations from self-supervised learning, models trained with CE loss, and models trained with CE+RMDNet loss.

\subsubsection{Self-supervised learning}\label{subsubsec2}

\begin{figure*}[!t]
\centering
\includegraphics[width=1.0\textwidth]{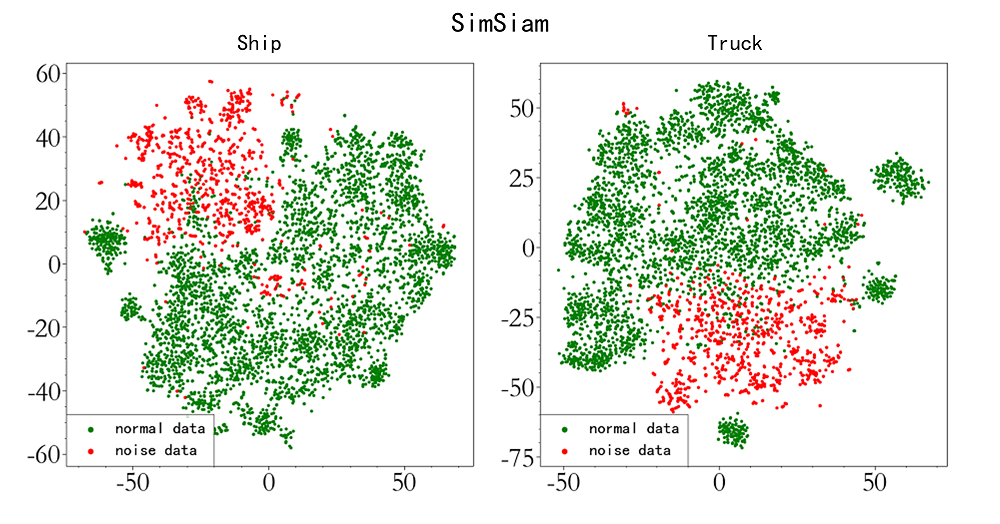}
\caption{SimSiam is trained on the CIFAR-10 dataset before extracting and visualising the representations for the classes Ship and Truck.}
\label{fig5}
\vspace{-0.3cm}
\end{figure*}

Self-supervised learning does not rely on data labels and therefore can effectively avoid the influence of noisy labels and extract reliable data representations. In this part, we adopt the SimSiam framework for the extraction of representations. Specific experiments are conducted on the CIFAR-10 dataset, which has a label noise type of symmetric noise and a noise rate of 0.2. We visualised the extracted representations from the final model, and Fig. \ref{fig5} shows the results of the visualisation of the extracted representations for the two categories of Ship and Truck in CIFAR-10.

As can be observed from Fig.  \ref{fig5}, the extracted features do not completely isolate the noisy data from the normal data completely due to certain limitations in the effect of self-supervised learning, but it is clear that in both figures, most of the red points are clustered together, while there are relatively few green points in the region where the red points are clustered. This indicates that although self-supervised learning did not completely isolate the noise, the extracted noisy data representations are somewhat different from the normal data representations.

\subsubsection{Cross entropy}\label{subsubsec2}

\begin{figure*}[!t]
\centering
\includegraphics[width=01.0\textwidth]{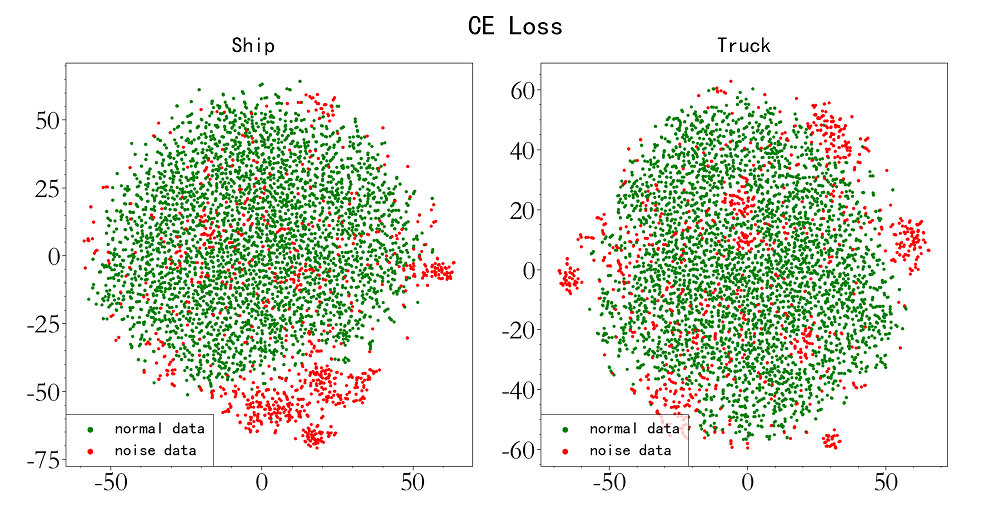}
\caption{After training on the CIFAR-10 dataset using CE as a loss function, two class representations, Ship and Truck, are extracted and visualised.}
\label{fig6}
\end{figure*}

In this experiment, we use CE as the loss function for the task network. The experiments were performed on the CIFAR-10 dataset with symmetric noise and a noise rate of 0.2. We visualised the extracted representations from the final model, and Fig. \ref{fig6} shows the results of the visualisation of the extracted representations for the two categories of Ship and Truck in CIFAR-10.

As can be observed from Fig. \ref{fig6}, when using only CE as the loss function, the noisy data represented in red is visually mixed with the normal data in green to a very high degree. This indicates a low level of differentiation between the extracted representations of the noisy data and the normal data, and this similarity may lead to an increase in the probability of the model incorrectly recognizing the noisy data as the normal data in the subsequent processing stages, especially in the classification process of the linear layer. This result reveals that if CE is used alone as the loss function, the model is difficult to distinguish the noisy data, resulting in poor model robustness.

\begin{figure*}[!t]
\centering
\includegraphics[width=01.0\textwidth]{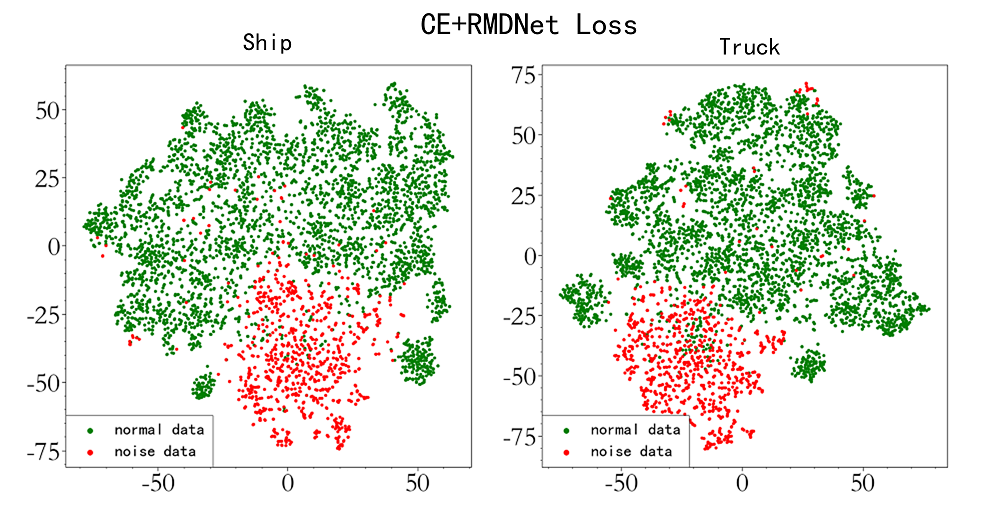}
\caption{After training on the CIFAR-10 dataset using CE+RMDNet as a loss function, two class representations, Ship and Truck, are extracted and visualised.}
\label{fig7}
\end{figure*}

\subsubsection{Cross entropy + RMDNet}\label{subsubsec2}

In this experiment, we adopted CE+RMDNet as the loss function, and made full use of the extracted representations from the task channel and the relational channel for the RMDNet loss computation, while setting the value of K to 1. The experiment was conducted on the CIFAR-10 dataset with symmetric noise and a noise rate of 0.2. Through training, We visualised the extracted representations from the final model, and Fig. \ref{fig7} shows the results of the visualisation of the extracted representations for the two categories of Ship and Truck in CIFAR-10.

It can be observed from Fig. \ref{fig7} that compared with the model using only cross entropy (CE) as the loss function, the model using CE+RMDNet loss function shows a significant improvement in the visualization graph: the red points (noisy data) are densely populated areas, and the green points (normal data) are relatively few. This phenomenon suggests that there is some difference between the representation of noisy data and normal data, which to some extent avoids the problem of the model misidentifying noisy data as normal data.

In summary, after using our method we are indeed able to distinguish, to some extent, the noisy data from the normal data, which further improves the accuracy of the training and enhances the robustness of the model.

\section{Conclusion}\label{sec5}

This paper proposes a relation modeling and distillation method for learning with noisy labels, called RMDNet. It first leverages self-supervised learning method to learn discriminative representations and model the relation between samples, which does not require guidance from labels, eliminating the interference of noise. Moreover, RMDNet follows knowledge distillation to calibrate the representation distribution of samples with noisy labels.  Experimental results show that RMDNet can combine multiple methods to enhancing representation learning, bringing them benefits. And RMDNet improves the robustness of the model in the inference phase.

Despite the impressive performance of RMDNet, there are still two directions worth exploring in future work.
Firstly, stronger self-supervised learning methods may bring performance gains to RMDNet, as they can model more accurate sample relationships. Secondly, pre-trained large language models can provide generalized knowledge, which may help RMDNet filter out some noise.

\section*{Competing interests}
No confict of interests in this paper that are directly or indirectly related to the work submitted for publication.

\section*{Ethical and informed consent for data used}
No ethical data in this paper.

\section*{Authors contribution statement}
Xiaming Chen: Investigation, Methodology, Writing—review, Formal analysis. Junlin Zhang: Investigation, Resources, Writing-original draft \& editing. Zhuang Qi: Supervision, Methodology, Writing—review \& editing. Xin Qi: Conceptualization, Supervision, Writing—review \& editing, Funding acquisition.

\section*{Data availability and access}
The data that support the findings of this study are openly available in http://www.cs.toronto.edu/-kriz/cifar-10-python.tar.gz and http://www.cs.toronto.edu/~kriz/cifar-100-python.tar.gz. And the code are available from the corresponding author on reasonable request after review.

\bibliography{sn-bibliography}

\begin{thebibliography}{51}
\providecommand{\natexlab}[1]{#1}
\providecommand{\url}[1]{{#1}}
\providecommand{\urlprefix}{URL }
\providecommand{\doi}[1]{\url{https://doi.org/#1}}
\providecommand{\eprint}[2][]{\url{#2}}
 \bibcommenthead

\bibitem[{Dang et~al(2024)Dang, Liu, Li, Xu, Wang, and Pan}]{dang2024multi}
Dang M, Liu G, Li H, et~al (2024) Multi-object behaviour recognition based on object detection cascaded image classification in classroom scenes. Applied Intelligence 54(6):4935--4951

\bibitem[{Xia and Ding(2024)}]{xia2024human}
Xia L, Ding X (2024) Human-object interaction detection based on cascade multi-scale transformer. Applied Intelligence pp 1--20

\bibitem[{Yang et~al(2022)Yang, Li, Jiang, Gong, Yuan, Zhao, and Yuan}]{yang2022focal}
Yang Z, Li Z, Jiang X, et~al (2022) Focal and global knowledge distillation for detectors. In: Proceedings of the IEEE/CVF Conference on Computer Vision and Pattern Recognition, pp 4643--4652

\bibitem[{Wu et~al(2021)Wu, Xu, Guo, Huang, Xu, Wang, and Li}]{9293326}
Wu F, Xu T, Guo J, et~al (2021) Deep siamese cross-residual learning for robust visual tracking. IEEE Internet of Things Journal 8(20):15216--15227. \doi{10.1109/JIOT.2020.3041052}

\bibitem[{Fernández-Sanjurjo et~al(2021)Fernández-Sanjurjo, Mucientes, and Brea}]{9344696}
Fernández-Sanjurjo M, Mucientes M, Brea VM (2021) Real-time multiple object visual tracking for embedded gpu systems. IEEE Internet of Things Journal 8(11):9177--9188. \doi{10.1109/JIOT.2021.3056239}

\bibitem[{Gan et~al(2023)Gan, Hu, Tan, and Du}]{gan2023tbnf}
Gan L, Hu L, Tan X, et~al (2023) Tbnf: A transformer-based noise filtering method for chinese long-form text matching. Applied Intelligence 53(19):22313--22327

\bibitem[{Wang et~al(2023)Wang, Chen, Obaidat, Kumari, Kumar, and Long}]{9474959}
Wang K, Chen CM, Obaidat MS, et~al (2023) Deep semantics sorting of voice-interaction-enabled industrial control system. IEEE Internet of Things Journal 10(4):2793--2801. \doi{10.1109/JIOT.2021.3093496}

\bibitem[{Niu et~al(2021)Niu, Xiao, Zhang, Zhang, Du, Huang, and Guizani}]{9220767}
Niu W, Xiao J, Zhang X, et~al (2021) Malware on internet of uavs detection combining string matching and fourier transformation. IEEE Internet of Things Journal 8(12):9905--9919. \doi{10.1109/JIOT.2020.3029970}

\bibitem[{Li et~al(2024)Li, Liu, Sun, Zhang, and Dou}]{li2024mask}
Li Q, Liu J, Sun Y, et~al (2024) On mask-based image set desensitization with recognition support. Applied Intelligence 54(1):886--898

\bibitem[{Naseem et~al(2023)Naseem, Rathore, Kumar, Gangopadhyay, and Jain}]{naseem2023approach}
Naseem S, Rathore SS, Kumar S, et~al (2023) An approach to occluded face recognition based on dynamic image-to-class warping using structural similarity index. Applied Intelligence 53(23):28501--28519

\bibitem[{Gao et~al(2021)Gao, Zhang, Yu, Lin, Wang, Yang, and Kong}]{9222066}
Gao P, Zhang H, Yu J, et~al (2021) Secure cloud-aided object recognition on hyperspectral remote sensing images. IEEE Internet of Things Journal 8(5):3287--3299. \doi{10.1109/JIOT.2020.3030813}

\bibitem[{Yang et~al(2024)Yang, Cui, Wang, and Wang}]{yang2024cross}
Yang S, Cui L, Wang L, et~al (2024) Cross-modal contrastive learning for multimodal sentiment recognition. Applied Intelligence pp 1--17

\bibitem[{Wang et~al(2019)Wang, Sahoo, Liu, Lim, and Hoi}]{wang2019learning}
Wang H, Sahoo D, Liu C, et~al (2019) Learning cross-modal embeddings with adversarial networks for cooking recipes and food images. In: Proceedings of the IEEE/CVF conference on computer vision and pattern recognition, pp 11572--11581

\bibitem[{Song et~al(2022)Song, Kim, Park, Shin, and Lee}]{song2022learning}
Song H, Kim M, Park D, et~al (2022) Learning from noisy labels with deep neural networks: A survey. IEEE transactions on neural networks and learning systems

\bibitem[{Smart and Carneiro(2023)}]{smart2023bootstrapping}
Smart B, Carneiro G (2023) Bootstrapping the relationship between images and their clean and noisy labels. In: Proceedings of the IEEE/CVF Winter Conference on Applications of Computer Vision, pp 5344--5354

\bibitem[{Wei et~al(2023)Wei, Feng, Sun, Wang, Guo, and Yin}]{wei2023fine}
Wei Q, Feng L, Sun H, et~al (2023) Fine-grained classification with noisy labels. In: Proceedings of the IEEE/CVF Conference on Computer Vision and Pattern Recognition, pp 11651--11660

\bibitem[{Lee et~al(2018)Lee, He, Zhang, and Yang}]{lee2018cleannet}
Lee KH, He X, Zhang L, et~al (2018) Cleannet: Transfer learning for scalable image classifier training with label noise. In: Proceedings of the IEEE conference on computer vision and pattern recognition, pp 5447--5456

\bibitem[{Li et~al(2022)Li, Xia, Ge, and Liu}]{li2022selective}
Li S, Xia X, Ge S, et~al (2022) Selective-supervised contrastive learning with noisy labels. In: Proceedings of the IEEE/CVF conference on computer vision and pattern recognition, pp 316--325

\bibitem[{Xiao et~al(2015)Xiao, Xia, Yang, Huang, and Wang}]{xiao2015learning}
Xiao T, Xia T, Yang Y, et~al (2015) Learning from massive noisy labeled data for image classification. In: Proceedings of the IEEE conference on computer vision and pattern recognition, pp 2691--2699

\bibitem[{Li et~al(2017)Li, Wang, Li, Agustsson, and Van~Gool}]{li2017webvision}
Li W, Wang L, Li W, et~al (2017) Webvision database: Visual learning and understanding from web data. arXiv preprint arXiv:170802862

\bibitem[{Song et~al(2019)Song, Kim, and Lee}]{song2019selfie}
Song H, Kim M, Lee JG (2019) Selfie: Refurbishing unclean samples for robust deep learning. In: International conference on machine learning, PMLR, pp 5907--5915

\bibitem[{Feng et~al(2021)Feng, Shu, Lin, Lv, Li, and An}]{feng2021can}
Feng L, Shu S, Lin Z, et~al (2021) Can cross entropy loss be robust to label noise? In: Proceedings of the twenty-ninth international conference on international joint conferences on artificial intelligence, pp 2206--2212

\bibitem[{Wang et~al(2019)Wang, Ma, Chen, Luo, Yi, and Bailey}]{wang2019symmetric}
Wang Y, Ma X, Chen Z, et~al (2019) Symmetric cross entropy for robust learning with noisy labels. In: Proceedings of the IEEE/CVF international conference on computer vision, pp 322--330

\bibitem[{Ghosh et~al(2017)Ghosh, Kumar, and Sastry}]{ghosh2017robust}
Ghosh A, Kumar H, Sastry PS (2017) Robust loss functions under label noise for deep neural networks. In: Proceedings of the AAAI conference on artificial intelligence

\bibitem[{Zhang and Sabuncu(2018)}]{zhang2018generalized}
Zhang Z, Sabuncu M (2018) Generalized cross entropy loss for training deep neural networks with noisy labels. Advances in neural information processing systems 31

\bibitem[{Tanaka et~al(2018)Tanaka, Ikami, Yamasaki, and Aizawa}]{tanaka2018joint}
Tanaka D, Ikami D, Yamasaki T, et~al (2018) Joint optimization framework for learning with noisy labels. In: Proceedings of the IEEE conference on computer vision and pattern recognition, pp 5552--5560

\bibitem[{Kim et~al(2019)Kim, Yim, Yun, and Kim}]{kim2019nlnl}
Kim Y, Yim J, Yun J, et~al (2019) Nlnl: Negative learning for noisy labels. In: Proceedings of the IEEE/CVF international conference on computer vision, pp 101--110

\bibitem[{Nguyen et~al(2019)Nguyen, Mummadi, Ngo, Nguyen, Beggel, and Brox}]{nguyen2019self}
Nguyen DT, Mummadi CK, Ngo TPN, et~al (2019) Self: Learning to filter noisy labels with self-ensembling. arXiv preprint arXiv:191001842

\bibitem[{Han et~al(2019)Han, Luo, and Wang}]{han2019deep}
Han J, Luo P, Wang X (2019) Deep self-learning from noisy labels. In: Proceedings of the IEEE/CVF international conference on computer vision, pp 5138--5147

\bibitem[{He et~al(2020)He, Fan, Wu, Xie, and Girshick}]{he2020momentum}
He K, Fan H, Wu Y, et~al (2020) Momentum contrast for unsupervised visual representation learning. In: Proceedings of the IEEE/CVF conference on computer vision and pattern recognition, pp 9729--9738

\bibitem[{Chen et~al(2020{\natexlab{a}})Chen, Kornblith, Norouzi, and Hinton}]{chen2020simple}
Chen T, Kornblith S, Norouzi M, et~al (2020{\natexlab{a}}) A simple framework for contrastive learning of visual representations. In: International conference on machine learning, PMLR, pp 1597--1607

\bibitem[{Chen et~al(2020{\natexlab{b}})Chen, Fan, Girshick, and He}]{chen2020improved}
Chen X, Fan H, Girshick R, et~al (2020{\natexlab{b}}) Improved baselines with momentum contrastive learning. arXiv preprint arXiv:200304297

\bibitem[{Chen et~al(2020{\natexlab{c}})Chen, Kornblith, Swersky, Norouzi, and Hinton}]{chen2020big}
Chen T, Kornblith S, Swersky K, et~al (2020{\natexlab{c}}) Big self-supervised models are strong semi-supervised learners. Advances in neural information processing systems 33:22243--22255

\bibitem[{Chen et~al(2021)Chen, Xie, and He}]{chen2021empirical}
Chen X, Xie S, He K (2021) An empirical study of training self-supervised vision transformers. In: Proceedings of the IEEE/CVF international conference on computer vision, pp 9640--9649

\bibitem[{Caron et~al(2021)Caron, Touvron, Misra, J{\'e}gou, Mairal, Bojanowski, and Joulin}]{caron2021emerging}
Caron M, Touvron H, Misra I, et~al (2021) Emerging properties in self-supervised vision transformers. In: Proceedings of the IEEE/CVF international conference on computer vision, pp 9650--9660

\bibitem[{Yi et~al(2022)Yi, Liu, She, McLeod, and Wang}]{yi2022learning}
Yi L, Liu S, She Q, et~al (2022) On learning contrastive representations for learning with noisy labels. In: Proceedings of the IEEE/CVF conference on computer vision and pattern recognition, pp 16682--16691

\bibitem[{Huang et~al(2023)Huang, Zhang, and Shan}]{huang2023twin}
Huang Z, Zhang J, Shan H (2023) Twin contrastive learning with noisy labels. In: Proceedings of the IEEE/CVF Conference on Computer Vision and Pattern Recognition, pp 11661--11670

\bibitem[{Karim et~al(2022)Karim, Rizve, Rahnavard, Mian, and Shah}]{karim2022unicon}
Karim N, Rizve MN, Rahnavard N, et~al (2022) Unicon: Combating label noise through uniform selection and contrastive learning. In: Proceedings of the IEEE/CVF Conference on Computer Vision and Pattern Recognition, pp 9676--9686

\bibitem[{Ma et~al(2022)Ma, Chen, and Akata}]{ma2022distilling}
Ma Y, Chen Y, Akata Z (2022) Distilling knowledge from self-supervised teacher by embedding graph alignment. arXiv preprint arXiv:221113264

\bibitem[{Chen et~al(2023)Chen, Yang, Peng, and Li}]{chen2023accelerating}
Chen Q, Yang H, Peng P, et~al (2023) Accelerating semi-supervised text classification by k-way projecting networks. IEEE Access 11:20298--20308

\bibitem[{Huang et~al(2022)Huang, You, Wang, Qian, and Xu}]{huang2022knowledge}
Huang T, You S, Wang F, et~al (2022) Knowledge distillation from a stronger teacher. Advances in Neural Information Processing Systems 35:33716--33727

\bibitem[{Yang et~al(2022)Yang, Zhou, An, Jiang, Xu, and Zhang}]{yang2022cross}
Yang C, Zhou H, An Z, et~al (2022) Cross-image relational knowledge distillation for semantic segmentation. In: Proceedings of the IEEE/CVF Conference on Computer Vision and Pattern Recognition, pp 12319--12328

\bibitem[{Chen and He(2021)}]{chen2021exploring}
Chen X, He K (2021) Exploring simple siamese representation learning. In: Proceedings of the IEEE/CVF conference on computer vision and pattern recognition, pp 15750--15758

\bibitem[{Zheng et~al(2020)Zheng, Wu, Goswami, Goswami, Metaxas, and Chen}]{zheng2020error}
Zheng S, Wu P, Goswami A, et~al (2020) Error-bounded correction of noisy labels. In: International Conference on Machine Learning, PMLR, pp 11447--11457

\bibitem[{Lyu and Tsang(2019)}]{lyu2019curriculum}
Lyu Y, Tsang IW (2019) Curriculum loss: Robust learning and generalization against label corruption. arXiv preprint arXiv:190510045

\bibitem[{Zhou et~al(2023)Zhou, Liu, Zhai, Jiang, and Ji}]{zhou2023asymmetric}
Zhou X, Liu X, Zhai D, et~al (2023) Asymmetric loss functions for noise-tolerant learning: Theory and applications. IEEE Transactions on Pattern Analysis and Machine Intelligence

\bibitem[{Qi et~al(2022)Qi, Wang, Chen, Wang, Meng, and Meng}]{qi2022clustering}
Qi Z, Wang Y, Chen Z, et~al (2022) Clustering-based curriculum construction for sample-balanced federated learning. In: CAAI International Conference on Artificial Intelligence, vol~10. Springer, pp 155--166

\bibitem[{Qi et~al(2023)Qi, Meng, Chen, Hu, Lin, and Meng}]{qi2023cross}
Qi Z, Meng L, Chen Z, et~al (2023) Cross-silo prototypical calibration for federated learning with non-iid data. In: Proceedings of the 31st ACM International Conference on Multimedia, vol~10. ACM, pp 3099--3107

\bibitem[{Qi et~al(2024{\natexlab{a}})Qi, Meng, He, Zhang, Wang, Qi, and Meng}]{qi2024cross}
Qi Z, Meng L, He W, et~al (2024{\natexlab{a}}) Cross-training with multi-view knowledge fusion for heterogenous federated learning. arXiv preprint arXiv:240520046

\bibitem[{Qi et~al(2024{\natexlab{b}})Qi, He, Meng, and Meng}]{qiattentive}
Qi Z, He W, Meng X, et~al (2024{\natexlab{b}}) Attentive modeling and distillation for out-of-distribution generalization of federated learning. In: 2024 IEEE International Conference on Multimedia and Expo (ICME), vol~10. IEEE, pp 648--653

\bibitem[{Liu et~al(2023)Liu, Qi, Chen, Meng, and Meng}]{liu2023cross}
Liu T, Qi Z, Chen Z, et~al (2023) Cross-training with prototypical distillation for improving the generalization of federated learning. In: 2023 IEEE International Conference on Multimedia and Expo (ICME), vol~10. IEEE, pp 648--653

\end{thebibliography}

\end{document}